\documentclass[runningheads]{llncs}

\usepackage[mobile]{eccv}

\usepackage{eccvabbrv}

\usepackage{graphicx}
\usepackage{booktabs}

\usepackage[accsupp]{axessibility}  

\usepackage{amsmath,amsfonts,bm}

\def\eqref#1{equation~\ref{#1}}

\def\1{\bm{1}}

\DeclareMathAlphabet{\mathsfit}{\encodingdefault}{\sfdefault}{m}{sl}
\SetMathAlphabet{\mathsfit}{bold}{\encodingdefault}{\sfdefault}{bx}{n}

\usepackage[utf8]{inputenc} 
\usepackage[T1]{fontenc}    
\usepackage{url}            
\usepackage{booktabs}       
\usepackage{amsfonts}       
\usepackage{nicefrac}       
\usepackage{microtype}      
\usepackage{xcolor}         

\usepackage{amsmath}
\usepackage{multirow}
\usepackage{tikz}
\usepackage{color}
\usepackage{pifont}
\usepackage{amssymb}
\usepackage{enumitem}
\usepackage{xspace}

\usepackage{graphicx}
\usepackage{caption} 
\usepackage{subcaption}
\usepackage{wrapfig}
\usepackage{tabulary}
\usepackage{multirow}
\usepackage{overpic}
\usepackage{xcolor}

\newcommand{\yes}{\ding{51}}
\newcommand{\gyes}{\textcolor{black!30}{\ding{51}}}
\newcommand{\no}{\ding{55}}

\newcolumntype{x}[1]{>{\centering\arraybackslash}m{#1pt}}
\newcolumntype{y}[1]{>{\arraybackslash}m{#1pt}}
\newcommand{\tablestyle}[2]{\setlength{\tabcolsep}{#1}\renewcommand{\arraystretch}{#2}\centering\footnotesize}

\usepackage[colorlinks,citecolor=eccvblue]{hyperref}

\usepackage{orcidlink}

\begin{document}

\title{ZeroI2V: Zero-Cost Adaptation of Pre-trained Transformers from Image to Video} 

\titlerunning{ZeroI2V}

\author{Xinhao Li\inst{1,2} \and
Yuhan Zhu\inst{1} \and
Limin Wang\inst{1,2}\thanks{Corresponding author.}}

\authorrunning{X.~Li et al.}

\institute{State Key Laboratory for Novel Software Technology, Nanjing University \and
Shanghai AI Laboratory \\
\email{xinhaoli00@outlook.com \quad zyuhan0812@gmail.com \quad lmwang@nju.edu.cn}
\url{https://github.com/MCG-NJU/ZeroI2V}}

\maketitle

\begin{abstract}
Adapting image models to the video domain has emerged as an efficient paradigm for solving video recognition tasks. Due to the huge number of parameters and effective transferability of image models, performing full fine-tuning is less efficient and even unnecessary. Thus, recent research is shifting its focus toward parameter-efficient image-to-video adaptation. However, these adaptation strategies inevitably introduce extra computational costs to deal with the domain gap and temporal modeling in videos. In this paper, we present a new adaptation paradigm (ZeroI2V) to transfer the image transformers to video recognition tasks (\ie, introduce zero extra cost to the original models during inference). To achieve this goal, we present two core designs. First, to capture the dynamics in videos and reduce the difficulty of image-to-video adaptation, we exploit the flexibility of self-attention and introduce spatial-temporal dual-headed attention (STDHA). This approach efficiently endows the image transformers with temporal modeling capability at zero extra parameters and computation. Second, to handle the domain gap between images and videos, we propose a linear adaption strategy that utilizes lightweight densely placed linear adapters to fully transfer the frozen image models to video recognition. Thanks to the customized linear design, all newly added adapters could be easily merged with the original modules through structural reparameterization after training, enabling zero extra cost during inference. Extensive experiments on representative fully-supervised and few-shot video recognition benchmarks showcase that ZeroI2V can match or even outperform previous state-of-the-art methods while enjoying superior parameter and inference efficiency.
\end{abstract}

\section{Introduction}
\label{sec:Introduction}
Adapting pre-trained foundation models such as BERT~\cite{bert} and GPT~\cite{radford2018improving, radford2019language, brown2020language} through efficient strategies has yielded excellent performance on downstream tasks in natural language understanding. This new paradigm is becoming popular in computer vision due to the available pre-trained image models such as CLIP~\cite{clip} and DINO~\cite{caron2021emerging, oquab2023dinov2}. These models could be easily adapted to downstream tasks through linear probes, fine-tuning, or even zero-shot recognition, exhibiting robustness and strong transfer capabilities similar to those of large-scale language models. Recently, {\em parameter-efficient transfer learning} (PETL)~\cite{vpt, adaptformer,nie2022pro,evl, stadapter,zhu2024awt} is becoming an efficient paradigm to adapt these large pre-trained models due to their huge numbers of parameters and high computational cost of full fine-tuning.

For video understanding, there exist several large pre-trained video models~\cite{videomae,videomaev2} from self-supervised learning, but these models are of high computational complexity due to the joint spatiotemporal attentions. Therefore, adapting pre-trained image models to the video domain through efficient strategies is still a practical solution to video recognition. In fact, the state-of-the-art video networks have long relied on the pre-trained image models by inflating the kernels~\cite{i3dk400,non-local,vivit,videoswin} or inserting plug-and-play temporal modules~\cite{tsn,tsm,tea,tam,tdn}. However, most of these methods necessitate full fine-tuning, which involves updating all the model parameters during training on video datasets. As the scale of pre-trained models increases, full fine-tuning becomes impractical due to the high training costs and the risk of overfitting or even catastrophic forgetting when the downstream data is limited. In addition, these methods often inevitably introduce extra costs to the adapted video models due to these newly added modules.

\begin{figure}[t]
    \centering
    \includegraphics[width=\textwidth]{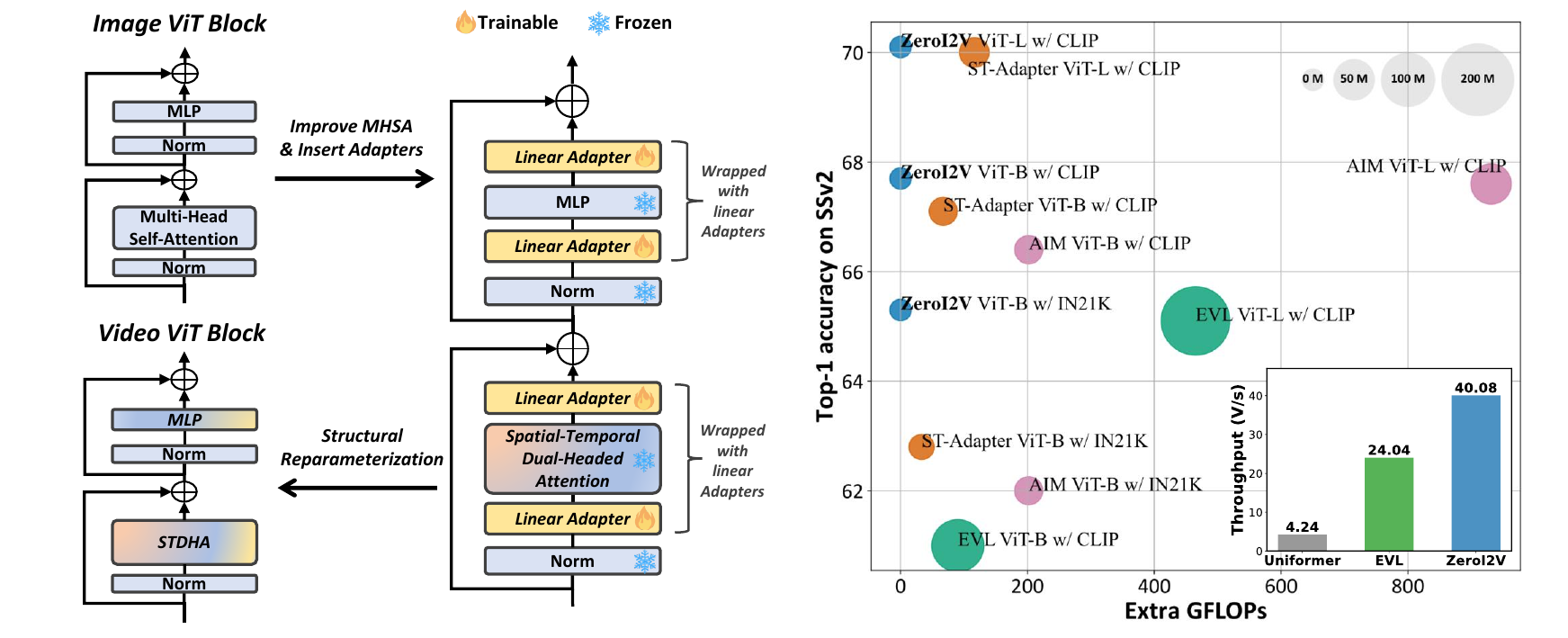}
    \caption{ \textbf{Left}: Our proposed image-to-video transfer learning method. \textbf{Right}: Comparison of PETL mehods on SSv2 validation set. For a more intuitive comparison, the views of the methods in the figure are all 8\(\times\)3 \(\times\)1. Two core techniques enable us to achieve superior performance on video tasks without introducing additional computation and parameters during inference.} 
    \label{fig:fig1}
    \vspace{-5mm}
\end{figure}

In this paper, we aim to present a new efficient paradigm of adapting image transformers to video downstream tasks with two main objectives. First, inspired by the PETL methods in NLP~\cite{houlsby2019parameter, lester2021power, lora, li2021prefix} and image understanding~\cite{vpt, adaptformer, nie2022pro}, we aim to devise a parameter-efficient transfer technique from image to video, which can effectively reduce the risk of over-fitting and greatly improve the training efficiency. Second, to overcome the issue of high computation in the adapted video models, we try to present a new adaptation method without introducing any extra computations to the final video models during inference. This zero extra inference cost adaptation would allow for more efficient deployment of transferred video models in real applications.

To achieve the above two objectives, we propose a novel transfer learning method (as shown in Figure \ref{fig:fig1}) that can utilize the off-the-shelf pre-trained image transformers to \textbf{\em achieve excellent performance on video tasks without additional parameters and computation during inference}. To be specific, for the temporal modeling required for video tasks, we transform multi-head self-attention into {\em spatio-temporal dual-head attention} (STDHA) by reassigning some heads to achieve temporal modeling at zero computation and zero parameters. For image-to-video transfer, we explore the strategy of using linear adapters to fully adapt the parameters of each part of the model and merge them with the frozen original parameters through structural reparameterization after training, thus achieving zero extra cost during inference.

To summarize, we make the following contributions:
    \textbf{1)} We propose a new approach for {\em parameter-efficient image-to-video transfer learning} that can achieve the efficient adaptation of transformers from image to video without introducing additional computation and parameters during inference.
    \textbf{2)} We introduce a novel attention mechanism named {\em Spatial-Temporal Dual-Headed Attention} (STDHA), which utilizes the flexibility of self-attention to achieve temporal modeling without introducing extra computation and parameters.
    \textbf{3)} To the best of our knowledge, we are the first to investigate the achievement of zero extra inference cost image-to-video adaptation through the utilization of a linear structure. We establish an empirical study by conducting extensive experiments with a diverse range of adaptation strategies.
    \textbf{4)} Our method achieves comparable or even better performance than state-of-the-art methods on popular fully-supervised and few-shot video recognition benchmarks while enjoying the advantage of parameter and inference efficiency.

\section{Related work}
\label{sec:Related_work}

\noindent\textbf{Pre-trained image transformers} The powerful scalability of ViT~\cite{vit} brings more possibilities to the pre-trained image model. In addition to the traditional supervised approach~\cite{vit, scalevit, swinv2}, recent works~\cite{moco, beit, caron2021emerging, mae, oquab2023dinov2} utilize self-supervised learning to effectively learn representations from unlabeled data. Moreover, several works~\cite{clip, li2022blip, tschannen2022image, cherti2022reproducible} adopt large-scale multi-modal data (\eg, text-image pairs) to learn visual representations with great transferability. Our proposed adaptation strategy can leverage these off-the-shelf pre-trained image transformers to achieve outstanding performance on video tasks.

\noindent\textbf{Video action recognition} is crucial for downstream tasks~\cite{dualdetr,pointtad}. Traditionally, state-of-the-art methods have long relied on image models. Previous works for action recognition can be classified into two categories: one is to extend the image model for spatial-temporal modeling by inflating weights and structures~\cite{i3dk400, slowfast, x3d, videoswin, uniformer, mvit, mvitv2}, while the other is to directly utilize the image model as the backbone and insert plug-and-play modules for temporal modeling~\cite{tsn, trn, tsm, tam, tdn}. Following the success of new training paradigms in image understanding, several works have attempted to learn transferable video representations via self-supervised learning~\cite{videomae, bevt, lu2023cmae, videomaev2} or multi-modal video-text pre-training~\cite{cpd20,actionclip, xclip, uniformerv2}. However, the above methods usually require full fine-tuning of the entire model or training from scratch, resulting in high training costs and additional computational overhead. In this work, we avoid the above problems by adapting the pre-trained image transformers to video tasks in an efficient manner.

\noindent\textbf{Parameter-efficient transfer learning} To address the issue of training inefficiency caused by the continuous growth of model size, Parameter-efficient transfer learning (PETL) is initially introduced in NLP ~\cite{houlsby2019parameter, adapterhub, adapterfusion, lester2021power, li2021prefix, lora, bitfit} and subsequently applied to vision tasks~\cite{vpt, lianscaling, he2022parameter, adaptformer, nie2022pro,provp,dpl,zhu2024awt}. These techniques aim to achieve comparable or even superior performance on other tasks by fine-tuning only a small subset of trainable parameters. Most PETL methods~\cite{vpt, adaptformer, he2022parameter, lianscaling,  zhang2022neural, nie2022pro,zhu2024awt} in vision domain are limited to transfer within the same modality (\eg, image-to-image or video-to-video). In contrast, our research focuses on image-to-video transfer learning. Despite progress made by recent studies~\cite{evl, stadapter, yangaim}, these methods require additional computation and parameters for temporal modeling of video tasks and image-to-video adaptation. For example, EVL~\cite{evl} incorporates an additional temporal transformer decoder, while ST-Adapter~\cite{stadapter} introduces additional adapters with depth-wise 3D convolution layers. Similarly, AIM~\cite{yangaim} adds extra adapters and necessitates an additional time attention calculation at each block. In contrast to previous works, our proposed method eschews the introduction of additional computation or parameters during inference, yet still achieves comparable or superior performance compared to previous methods.

\section{Methodology}
\label{sec:Methodology}

In this section, we first briefly revisit the basic block of ViT (Sec. \ref{subsec:Preliminary}), and then discuss how to utilize the flexibility of self-attention to achieve temporal modeling without introducing additional computation and parameters (Sec. \ref{subsec:zerotemporal}). Finally, we explain how we implement zero-cost image-to-video adaptation with a serial linear structure (Sec. \ref{subsec:zeroadaptation}).

\subsection{Preliminary}
\label{subsec:Preliminary}

The original ViT~\cite{vit} block consists of two network layers: multi-head self-attention (MHSA) and multi-layer perceptron (MLP). As shown in Figure \ref{fig:fig1}, a ViT block consists of MHSA and MLP connected in series in a residual structure:
\begin{align}
    z_l &= x_l + \text{MHSA}(\text{LN}(x_l)), \\
x_{l+1} &= z_l+\text{MLP}(\text{LN}(z_l)),
\end{align}
where LN denotes layer normalization~\cite{layernorm} and \(x_l\) represents the input to the \(l\)-th ViT block. We review their specific implementation details. For the sake of simplicity, we ignore the bias and denote \(X \in \mathbb{R}^{n\times d}\) as input of MHSA and MLP. 

MHSA first performs three different linear projections \( W^Q_{\text{attn}}, W^K_{\text{attn}}, W^V_{\text{attn}} \in \mathbb{R}^{d\times d}\) on the input \(X \) to obtain the query  \( Q\) and key-value pairs \( K, V\). These are then evenly divided into \(h\) heads by channel. Each head independently performs the scaled dot-product attention calculation. Finally, the heads are concatenated by channel and then a linear projection \( W^O_{\text{attn}} \in \mathbb{R}^{d\times d}\) is performed to obtain the final calculation result:
\begin{align} 
    Q, K, V &= XW^Q_{\text{attn}}, XW^K_{\text{attn}}, XW^V_{\text{attn}}, \\
    \text{head}_i &= \text{Attention}(Q_i, K_i, V_i), \\
    \text{MHSA}(X) &= \text{Concat}(\text{head}_1, \cdots, \text{head}_h) W^O_{\text{attn}}.
\end{align}
MLP involves two linear projections \(W^{\text{up}}_{\text{mlp}} \in \mathbb{R}^{d\times d'}, W^{\text{down}}_{\text{mlp}}\in \mathbb{R}^{d'\times d}, d'>d\) and one non-linear activation function \(\sigma \):
\begin{align}
    \text{MLP}(X) &= \sigma(XW^{\text{up}}_{\text{mlp}}) W^{\text{down}}_{\text{mlp}}.
\end{align}

\begin{figure}[t]
    \centering
    \begin{overpic}[width=0.9\linewidth]{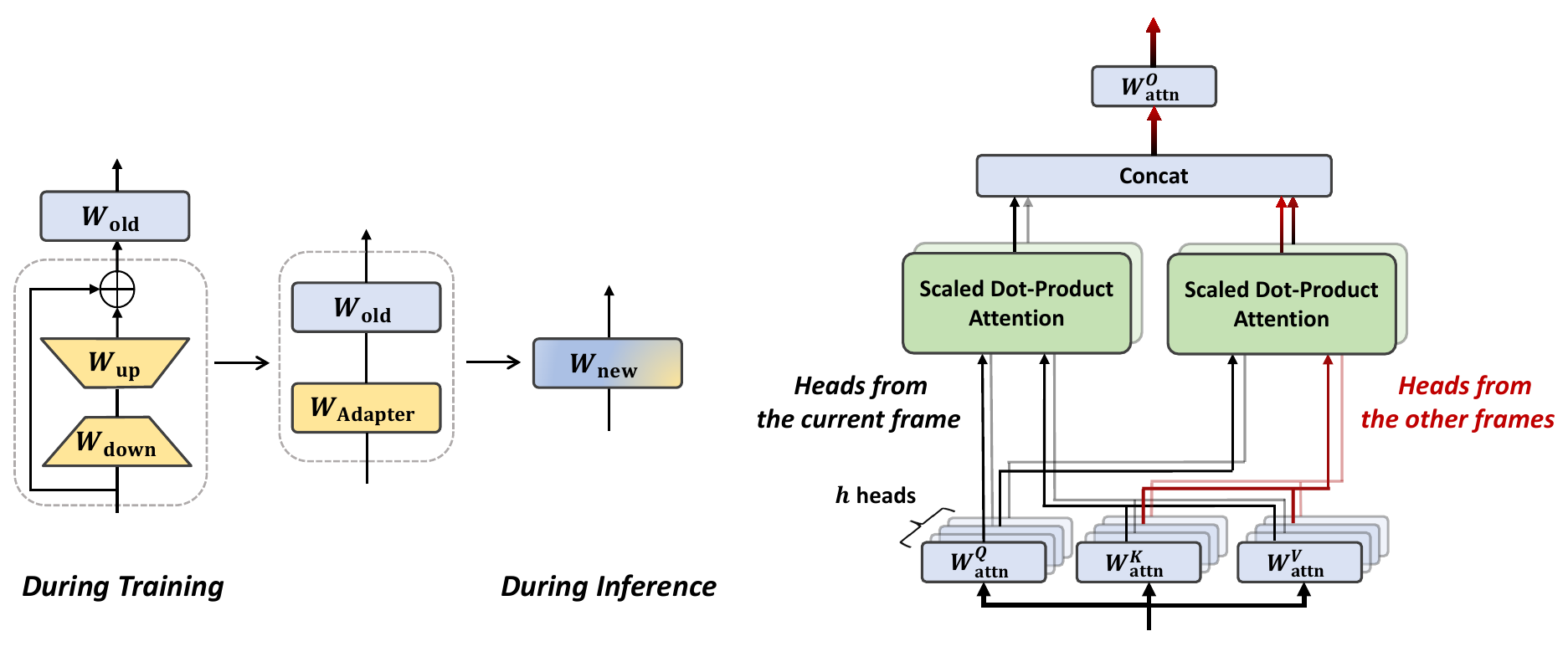}
        \put(-5,-0.4){ \footnotesize{(a) Layer merging via reparameterization}}
        \put(52,-0.4){ \footnotesize{(b) Spatial-temporal dual-headed attention}}
    \end{overpic}
    \caption{\textbf{Illustration of the proposed linear adaptation and STDHA.}}
    \label{fig:method}
    \vspace{-4mm}
\end{figure}

\subsection{Zero-Cost temporal modeling}
\label{subsec:zerotemporal}

Applying image models to video tasks often requires the incorporation of additional modules for temporal modeling, which not only introduces additional parameters and computation, but also results in additional training costs. In this work, we address temporal modeling from three key perspectives: (1) Capability of capturing the temporal dynamics. (2) Reducing the difficulty of image-to-video adaptation. (3) Minimizing the introduction of additional computation and parameters compared to the original model. \cite{head} suggests that most heads are redundant given the rest of the model. Inspired by this, we attempt to reassign some heads as \textit{temporal heads} in the multi-head attention to perform temporal modeling tasks, while the remaining heads continue to perform spatial modeling tasks as \textit{spatial heads}, thereby achieving efficient spatial-temporal modeling.



\noindent\textbf{Spatial-temporal dual-headed attention (STDHA)} As shown in Figure \ref{fig:method}b, consider an input sequence \(X =\{x_1, x_2, \cdots, x_T\}\) where \( x_t \in \mathbb{R}^{n \times d}\). Let the query and key-value pairs obtained after the linear projection of the \(x_t\) be \(Q^t,K^t,V^t \in \mathbb{R}^{n\times d}\). We divide the \(h\) heads of the MHSA into two groups of size \(h-k\) and  \(k\). One group of heads queries the key-value pairs at the current time \(t\) to perform \textit{spatial modeling}, while the other group of heads queries the key-value pairs at other times \(t+\Delta t_i\) to perform \textit{temporal modeling}. Finally, the information from the two groups of heads is aggregated by a linear projection to perform \textit{spatial-temporal modeling}:
\begin{align}
    \text{S-head}_i &= \text{Attention}(Q_i^t, K_i^t, V_i^t), \\
\text{T-head}_i &= \text{Attention}(Q_i^t, K_i^{t+\Delta t_i}, V_i^{t+\Delta t_i}) (\Delta t_i \ne 0), \\
    \text{STDHA}(X) &= \text{Concat}(\text{T-head}_1, \cdots,\text{T-head}_k, \text{S-head}_{k+1} \cdots \text{S-head}_h) W^O_{\text{attn}},
\end{align}
where \(\Delta t_i\) represents the time offset of the key-value pair of the \(i\)-th head. We did not directly use temporal attention or temporal convolution for the temporal modeling like previous works~\cite{evl, stadapter, yangaim}. Instead, we design a more efficient spatiotemporal modeling operator by decoupling spatial modeling and temporal modeling to different heads:

\begin{itemize}
    \item For the spatial head, it still only needs to complete the spatial modeling task as the original image transformer, which reduces the difficulty of achieving image-to-video adaptation.

    \item For the temporal head, it actually implements the inter-frame attention mechanism with frames at different times. ~\cite{motion} have demonstrated the effectiveness of an inter-frame attention mechanism for modeling motion information, which is crucial for action recognition tasks. In addition, as shown in Table \ref{tab:effect_HR_dist}, we can achieve both short-distance and long-distance modeling by controlling the \(\Delta t_i\) of the temporal head, which enables us to achieve enhanced temporal modeling capabilities.


\end{itemize}

\noindent\textbf{Comparison with other zero-cost operators} There have been several previous attempts~\cite{xvit, tokenshift, tps} to use image transformers to achieve efficient temporal modeling at zero parameters and zero computation. For example, ~\cite{xvit} achieves approximations to full space-time attention by mixing tokens from adjacent frames. ~\cite{tokenshift} performs temporal modeling by using channel shift on the cls tokens of different frames. ~\cite{tps} mixes information from adjacent frames using temporal patch shift and temporal channel shift before MHSA. However, these methods do not take advantage of the inherent characteristics of the transformer structure. 
By decoupling the learning of spatial and temporal information with head relocation, STDHA maintains the \textbf{\textit{purity of key-value pair information}} within the same head, thereby achieving better spatial-temporal information learning than other zero-cost temporal modules. And STDHA \textbf{\textit{simultaneously captures both short-range and long-range dependencies}}, rather than being limited to adjacent frames. As shown in Table \ref{tab:stdha_ablation}, these two key distinctions enable our STDHA to achieve superior spatial-temporal modeling.

\subsection{Zero Extra Inference Cost image-to-video adaptation}
\label{subsec:zeroadaptation}

Inspired by LoRA~\cite{lora}, we can fine-tune the model using a linear structure and then merge it with the original model during inference. However, to deal with the domain gap between images and videos, previous works~\cite{evl,yangaim,stadapter} often use nonlinear structures to achieve stronger transfer capabilities. Therefore, we need to further consider how to achieve effective image-to-video transfer using only a linear structure.

\noindent\textbf{Layer merging via structural reparameterization} Let \(W_{\text{old}}\) represent the frozen weights of the original model, and \(W_{\text{new}}\) represent the new trainable weights. Reviewing the structure of LoRA, it uses a low-rank decomposition matrix \(W_{\text{LoRA}}\) parallel to the original weights:
\begin{align}
    W_{\text{new}} = W_{\text{LoRA}} + W_{\text{old}} 
 = W_{\text{up}}W_{\text{down}}+  W_{\text{old}}.
\end{align}
 In this work, we use a serial linear structure called \textbf{\textit{Linear Adapter}} to fine-tune the original parameters. As shown in Figure \ref{fig:method}a, we use structural reparameterization to perform layer merging after training:
\begin{align}
    W_{\text{new}} = W_{\text{Adapter}}W_{\text{old}} 
    = (I+W_{\text{up}}W_{\text{down}}) W_{\text{old}},
\end{align}
where \(I\) is the identity matrix, \(W_{\text{up}} \in \mathbb{R}^{m\times k}, W_{\text{down}}\in \mathbb{R}^{ k \times n}\), bottleneck width \(k \ll \min(m, n)\). As seen in Table \ref{tab:petl_compare}, compared to parallel structures, serial structures can be more flexibly inserted into the network structure (\eg, for non-square matrices, under the same bottleneck dimension, using LoRA requires a larger number of parameters compared to Linear Adapter), which endows it with better transfer capabilities. 

\noindent\textbf{Full adaptation with densely placed linear adapters} By observing the structure of MHSA and MLP, we can see that all their trainable parameters concentrate on the linear projections at both ends of the structure. Therefore, fine-tuning the model essentially updates these linear projections. Previous works~\cite{yangaim,stadapter} often selectively tune part of the parameters (\eg, placing only an adapter before MHSA) instead of tuning all parameters to avoid excessive additional computational and parameter costs, while we can achieve zero-cost \textit{\textbf{full adaptation}} by tuning all parameters through wrapping MHSA and MLP with linear adapters. Table \ref{tab:petl_compare} shows that full adaptation enables us to achieve excellent image-to-video transfer performance with a linear structure, compensating for the performance degradation caused by the removal of nonlinearity.

\section{Experiments}
\label{sec:Experiments}

\subsection{Experiments setup}

We evaluate our method on five widely-used video recognition benchmarks: two large-scale datasets, namely Kinetics-400 (K400)~\cite{i3dk400} and Something-Something V2 (SSv2)~\cite{sth}, in addition to three smaller-scale datasets, UCF101~\cite{ucf101}, HMDB51~\cite{hmdb51} and Diving48~\cite{diving48}. We also evaluate our
method on action detection dataset AVA~\cite{ava}. This diverse dataset selection allows for a comprehensive evaluation of our model across various scales and domains. The specific model configuration and training strategy can be found in the supplementary. For most main experiments, we use ViT-B and ViT-L pre-trained by CLIP~\cite{clip} as our backbone models.



\subsection{Ablation study}

To validate the effectiveness of our method on image-to-video transfer and temporal modeling, we first conduct ablation experiments on the SSv2 dataset. All ablation experiments were performed using ViT-B/16 with 8 input frames unless specified.

\begin{table}[t]
\centering

\caption{\textbf{Ablation study on STDHA.} Most of the symbols in the table have been declared in the methodology section~\ref{sec:Methodology}. (a)\(R_c\) denotes channel change ratio, "Shift’ refers to temporal channel shift, while "HR" denotes head relocation as used by STDHA. (b) We use a multiset to represent the time offsets of different heads (\eg, "\(1 \cdot 2\)" means that there are 2 heads with \(\Delta t =1\)). When \(\Delta t\)=0, it represents a spatial head. (c) "Temporal RF" refers to temporal receptive field of a single STDHA.} \vspace{-2mm}
\label{tab:stdha_ablation}
\subfloat[Comparison of temporal modeling methods, ]{%
\tablestyle{4.8pt}{1}
\label{tab:shiftcompare}
\resizebox{0.43\textwidth}{!}{
\begin{tabular}{@{}x{30}x{80}x{40}@{}}
\toprule
\(R_c\) & Method & Top-1 \\
\toprule 
\multirow{5}{*}{1/6} & [cls] token shift     & 61.4 \\
& Shift \(QKV\) & 64.5 \\
& Shift \(KV\) & 64.6 \\
& HR \(QKV\) & 64.8 \\
& HR \(KV\) (STDHA) & \textbf{66.0} \\
\midrule
\multirow{2}{*}{1/4} & Shift KV & 64.0 \\
& HR \(KV\) (STDHA) & 65.8 \\
\bottomrule
\end{tabular}}
}\hfill
\vspace{-4.5mm}
\subfloat[Effect of the number of temporal heads]{%
\tablestyle{4.8pt}{1}
\label{tab:effect_HR_num}
\resizebox{0.5\textwidth}{!}{
\begin{tabular}{@{}x{40}x{85}x{20}x{30}@{}}
    \toprule
    Backbone & \(\Delta t\) of heads & \(k\) & Top-1 \\
    \midrule
    \multirow{4}{10mm}{ViT-B (\(h\)=12)} & 
    \(\{1\cdot\frac{1}{2}, -1\cdot\frac{1}{2}, 0\cdot11\} \) & 1    &  64.8    \\
  & \(\{1\cdot1, -1\cdot1, 0\cdot10\} \)& 2 & \textbf{66.0}    \\
  & \(\{1\cdot2, -1\cdot2, 0\cdot8\} \) & 4 & 65.6    \\
  & \(\{1\cdot3, -1\cdot3, 0\cdot6\} \) & 6 & 65.6    \\
  \midrule
\multirow{3}{10 mm}{ViT-L (\(h\)=16)}              & 
    \(\{1\cdot1, -1\cdot1, 0\cdot14\} \) & 2  & 67.7    \\
  & \(\{1\cdot2, -1\cdot2, 0\cdot12\} \) & 4  & \textbf{68.5}    \\
  & \(\{1\cdot3, -1\cdot3, 0\cdot10\} \) & 6  &  68.3   \\
  \bottomrule
  \end{tabular}}}\hfill
\subfloat[Effect of temporal receptive field at at different input lengths.]{%
    \tablestyle{4.8pt}{0.3}
    \label{tab:effect_HR_dist}
    \resizebox{0.9\textwidth}{!}{
        \begin{tabular}{@{}x{30}lx{60}x{30}@{}}
          \toprule
          Frames  & \(\Delta t\) of heads & Temporal RF & Top-1 \\
         \midrule
         \multirow{4}{*}{8} & 
         \(\{ 1\cdot1, 0\cdot11\} \)     & 2 &  64.7  \\
         &\(\{1\cdot1, -1\cdot1, 0\cdot10\} \)  & 3 & \textbf{66.0}    \\
         & \(\{1\cdot1, -1\cdot1, 2\cdot1, 0\cdot9\} \) & 4  &  65.5   \\
         & \(\{1\cdot1, -1\cdot1, 2\cdot1, -2\cdot1, 0\cdot8\} \) & 5  &  65.7   \\
        \midrule
      \multirow{5}{*}{16}              & 
      \(\{1\cdot1, -1\cdot1, 0\cdot10\} \)  & 3  & 67.2  \\
      & \(\{1\cdot1, -1\cdot1, 2\cdot1, 0\cdot9\} \) & 4 &  67.3   \\
      & \(\{1\cdot1, -1\cdot1, 2\cdot1, -2\cdot1, 0\cdot8\} \) & 5 & \textbf{67.8} \\
      & \(\{1\cdot1, -1\cdot1, 2\cdot1, -2\cdot1, 3\cdot1, 0\cdot7\} \) & 6 & 67.6 \\
      & \(\{1\cdot1, -1\cdot1, 2\cdot1, -2\cdot1, 3\cdot1, -3\cdot1, 0\cdot6\} \) & 7 & 67.3 \\
        \midrule
        \multirow{6}{*}{32}              & 
          \(\{1\cdot1, -1\cdot1, 0\cdot10\} \)   & 3 &  67.3   \\
             & \(\{1\cdot1, -1\cdot1, 2\cdot1, 0\cdot9\} \) & 4 & 67.8 \\
        & \(\{1\cdot1, -1\cdot1, 2\cdot1, -2\cdot1, 0\cdot8\} \) & 5 & 68.5 \\
        & \(\{1\cdot1, -1\cdot1, 2\cdot1, -2\cdot1, 3\cdot1, 0\cdot7\} \) & 6 & \textbf{68.6} \\
        & \(\{1\cdot1, -1\cdot1, 2\cdot1, -2\cdot1, 3\cdot1, -3\cdot1, 0\cdot6\} \) & 7 & 68.4  \\
        & \(\{1\cdot1, -1\cdot1, 2\cdot1, -2\cdot1, 3\cdot1, -3\cdot1, 4\cdot1, 0\cdot5\} \) & 8 & 68.2 \\
        \bottomrule

        \end{tabular}}}
        \vspace{-6mm}
\end{table}

\noindent\textbf{Effectiveness of STDHA} Table \ref{tab:shiftcompare} compares STDHA with other zero-cost temporal modeling methods. The [cls] token shift is implemented according to the original paper~\cite{tokenshift},  with [cls] token shift performed before MHSA and MLP. The temporal channel shift operation refers to TPS~\cite{tps}, which shifts a portion of the channels for each head. It can be seen that STDHA significantly outperforms other methods at the same channel change ratio, demonstrating the importance of preserving the purity of information within each head.

\noindent\textbf{Effect of the number of temporal heads and temporal receptive field} We examined the influence of the number of temporal heads and the temporal receptive field in ViT-B and ViT-L. Our findings, detailed in Tables \ref{tab:effect_HR_num} and \ref{tab:effect_HR_dist}, suggest that the optimal proportion of temporal heads in ViT lies between 1/6 and 1/4. For the temporal receptive field, our results indicate that for 8-frame inputs, a field of 3 is sufficient, while for longer inputs (16/32 frames), performance improves with an increase in the field from 3, saturating at around 5 or 6. Hence, we employ different STDHA configurations based on input length.

\begin{wraptable}[7]{r}{4.4cm}
\vspace{-10mm}
        \centering
        \caption{\textbf{Effect of bottleneck ratio of linear adapters.}}\label{tab:bottleneck}
        \renewcommand{\arraystretch}{0.7}
        \resizebox{0.32\columnwidth}{!}{
        \begin{tabular}{@{}x{40}x{40}x{40}@{}}
            \toprule
           Ratio & Tunable Param(M) & Top-1 \\
           \midrule
           0.0625 & 3 & 64.2 \\
           0.125 & 7 & 65.0 \\
           0.25 & 14 & \textbf{66.0} \\
           0.5 & 28 & 65.8 \\
            \bottomrule
        \end{tabular}}
\end{wraptable}

\noindent\textbf{Effect of bottleneck ratio} We set the bottleneck dimension equal to the product of the ViT width and the bottleneck ratio. As shown in Table \ref{tab:bottleneck}, a bottleneck ratio of 0.25 achieves a good trade-off between performance and the number of parameters. Therefore, we choose 0.25 as the bottleneck ratio for all subsequent experiments.

\begin{table}[t]
\tablestyle{2pt}{0.95}
    \centering
     \setlength\tabcolsep{2pt}
    \caption{\textbf{Comparison of adaption strategies.} "Width" refers to the bottleneck width of LoRA/Adapter. "Tunable Params" refers to extra trainable parameters besides the parameters of the ViT backbone and linear classifier. “\yes" and ”\no" indicate whether the corresponding weights have undergone fine-tuning, and "\gyes" indicates that \(W_{\text{attn}}^Q, W_{\text{attn}}^K\) and \(W_{\text{attn}}^V\) share the same adapter. "Latency" refers to inference latency with 3 samples. All results are obtained using a same V100-32G with PyTorch-builtin mixed precision.}
    \vspace{-3mm}
    \label{tab:petl_compare}
    \resizebox{\linewidth}{!}{
    \begin{tabular}{c c c c c c c c c c  c}
        \toprule
\multirow{2}{10mm}{Method}  & \multicolumn{6}{c}{Weights of ViT block} 
 & \multirow{2}{15mm}{\centering {Tunable Params(M)}} & \multirow{2}{15.5mm}{\centering Bottleneck Width} & \multirow{2}{8.5mm}{\centering \small Latency (ms)} & 
\multirow{2}{9mm}{\centering \small  SSv2 Top-1} \\
        \cmidrule(r){2-7}
                         & \(W_{\text{attn}}^Q\) & \(W_{\text{attn}}^K\) & \(W_{\text{attn}}^V\) & \(W_{\text{attn}}^O\) & \(W_{\text{mlp}}^{\text{up}}\) & \(W_{\text{mlp}}^{\text{down}}\)  &    &  \\
        
         \midrule
        Full Fine-tuning & \yes  & \yes & \yes   & \yes & \yes & \yes & 86 & -  & 28.9 & 63.2  \\
        \midrule
        Linear Probe &   \no  &  \no  &   \no &  \no  &  \no  &  \no &  0 & - &   28.9 &20.0 \\
             \midrule
        Only tuning temporal head &   \yes  &  \yes  &   \yes  &  \yes  &  \no  &  \no &  4.6 &  - &  28.9   &  59.6 \\
        \midrule
         \multirow{2}{*}{ST-Adapter~\cite{stadapter}} & \gyes  & \gyes & \gyes & \yes  & \yes & \yes & 14 & 192 &  41.0 & 66.2 \\
       & \gyes  & \gyes & \gyes & \no  & \yes & \no & 14 & 384 & 38.8 &65.8 \\
        \midrule
       \multirow{6}{*}{LoRA~\cite{lora}} & \yes  &   \no & \yes &   \no &  \no  &  \no & 7 & 192  &\multirow{7}{*}{28.9} & 64.2 \\

         & \yes  & \yes & \yes &   \yes &  \no  &  \no & 14  & 192 &   & 65.0 \\
            & \yes  &   \no & \yes &   \no &  \yes  &  \yes & 25 & 192 &   & 64.3  \\
          & \yes  &   \no & \yes &   \no &  \yes  &  \yes & 17 & 128 &   & 65.6  \\
         
        & \yes  &   \yes & \yes &   \yes &  \yes  &  \yes & 32 & 192 &  & 65.0  \\
          & \yes  &   \yes & \yes &   \yes &  \yes  &  \yes & 21 & 128 &  & 65.5  \\
          
        \midrule
        \multirow{4}{*}{Adapter w/ GELU}
        & \gyes  & \gyes & \gyes    & \yes & \yes & \yes & 7 & 96 &   37.3 & 65.6  \\
        & \gyes & \gyes & \gyes   &   \no & \yes &  \no  & 7 & 192 &  34.9 & 64.6 \\
        & \gyes  & \gyes &  \gyes  & \yes & \yes &   \no & 10 & 192 &  36.3 & 66.1  \\
       & \gyes  & \gyes & \gyes    & \yes & \yes & \yes & 14 & 192 &  38.4 & 66.1  \\
       \midrule
      
       \multirow{6}{*}{Linear Adapter (Ours)}  
       & \gyes  & \gyes & \gyes   & \yes & \yes &  \yes  & 7  & 96 &  \multirow{6}{*}{28.9} & 65.0 \\
       & \gyes & \gyes   & \gyes &   \no & \yes & \no  & 7 & 192  &  & 64.4  \\
       
      & \gyes  & \gyes & \gyes   & \yes & \yes &  \no  & 10 & 192 &  & 65.2  \\
      
      & \gyes  & \gyes & \gyes   & \yes & \yes & \yes & 14 & 192 &   & 66.0  \\
      & \yes  & \yes & \yes   & \yes & \yes & \yes & 20 & 192 &  & \textbf{66.3}  \\
       & \yes  & \yes & \yes   & \yes & \yes & \yes & 14 & 128 &  & 66.2  \\        \bottomrule
    \end{tabular}
    }\vspace{-5mm}
\end{table}

\noindent\textbf{Comparison of adaptation strategies} In Table \ref{tab:petl_compare}, we compare the image-to-video transfer ability of our method with a diverse range of adaptation methods. For a fair comparison, we all use STDHA with the same setting to provide temporal modeling capabilities. From the results, we can observe that:
\begin{itemize}
    \item Even with minimal parameters being fine-tuned, our Linear Adapter significantly outperforms full fine-tuning (66.3 vs 63.2).  Despite updating the fewest parameters, the linear probe performs poorly in image-to-video transfer. 

    \item Tuning only the temporal head achieves about 95\% of the full fine-tuning performance, suggesting that extensive fine-tuning of the spatial head may not be necessary to attain satisfactory transfer performance due to the decoupling of spatial and temporal modeling reduces the difficulty of adaptation.

    \item Our \textit{Full Adaptation} strategy is not only effective for linear adapters, but also for non-linear adapters such as the ST-Adapter and GELU Adapter. It not only enhances their adaptation performance, but also eliminates the performance gap between linear and non-linear structures. 
    

    \item Due to the inflexibility of the parallel structure, for non-square matrices like \(W_{\text{mlp}}\), LoRA requires more parameters under the same bottleneck width. It needs to decrease the bottleneck width of the low-rank matrix to align it with the number of parameters of the linear adapter. However, this reduction in bottleneck width can limit its adaptation ability, ultimately leading to results that are significantly worse than those of the Linear Adapter.
\end{itemize}

\begin{table}[t] 
    \centering
    \caption{\textbf{Results on Kinetics-400 validation set}. Views = \#frames \(\times\) \#spatial crops \(\times\) \#temporal clips. “GFLOPs” means \(10^9\) FLOPs, "M" means \(10^6\). “Extra GLOPs” refers to the extra computation added to the original ViT under the same number of views. "New Params" refers to additional parameters 
 during inference besides the parameters of the original ViT backbone and linear classifier.}\vspace{-3mm}
    \tablestyle{4.8pt}{1.0}
    \resizebox{\linewidth}{!}{
\begin{tabular}{@{}y{140}x{30}x{32}x{28}x{28}x{28}x{28}x{22}x{22}@{}}
\toprule
Methods & Pretrain & Views & GFLOPs & Extra GFLOPs & Param(M) & New Param(M) & Top-1 & Top-5\\
\toprule
\multicolumn{7}{@{}l}{\it Methods with full fine-tuning} \\



UniFormer-B~\cite{uniformer} & IN1K & 32\(\times\)3\(\times\)4 & 3108 & -  & 50 &- &  83.0 & 95.4\\

TimeSformer-L~\cite{timesformer} & IN21K & 96\(\times\)3\(\times\)1 & 7140 & - & 121 & - &  80.7 & 94.7  \\


VideoSwin-L~\cite{videoswin} & IN21K & 32\(\times\)3\(\times\)4 & 7248 & - & 197 & - & 83.1 & 95.9  \\
MViTv2-L(\(\uparrow\)312)~\cite{mvitv2} & IN21K & 40\(\times\)5\(\times\)3 & 42420 & - & 218 & - & 86.1 & 97.0  \\

ViViT-L/16x2 FE~\cite{vivit} & JFT & 32\(\times\)3\(\times\)1 & 11940 & - & 311 & - & 83.5 & 94.3 \\
MTV-L~\cite{multiview} & JFT  &32\(\times\)3\(\times\)4  & 18050 & -  & 876 & - & 84.3 & 96.3 \\

 ViT-B/16~\cite{stadapter} & CLIP & 8\(\times\)1\(\times\)3 & 422 & 0 & 86 & 0 & 81.0  & 95.5 \\
 ActionCLIP-B/16~\cite{actionclip} & CLIP & 32\(\times\)3\(\times\)10 & 16893 & 13  & 142 &  56 &  83.8 & 97.1 \\
 X-CLIP ViT-L/14~\cite{xclip} & CLIP & 8\(\times\)3\(\times\)4 & 7896  & 107 & 420 &  116 &   87.1 & 97.6 \\

  Text4Vis ViT-L/14~\cite{text2vis} & CLIP & 32\(\times\)3\(\times\)4 & 19944  & - & 347 & 43  &  87.1 &  97.4 \\
\midrule
\multicolumn{7}{@{}l}{\it Methods with PETL} \\

VideoPrompt ViT-B/16~\cite{videoprompt} & CLIP & 16\(\times\)5\(\times\)1 & - & - & - & - & 76.9  & 93.5 \\
 ST-Adapter ViT-B/16~\cite{stadapter}  & IN21K & 8\(\times\)1\(\times\)3 & 455 & 33  & 93 & 7 &  76.6 & - \\
ST-Adapter ViT-L/14\cite{stadapter} & CLIP & 32\(\times\)1\(\times\)3 & 8248 & 322 & 19 &  87.2 & 97.6 \\

  EVL ViT-B/16~\cite{evl} & IN21K & 8\(\times\)1\(\times\)3 & 454 & 32 & 115 & 29 & 75.4   &  -\\
 EVL ViT-L/14~\cite{evl} & CLIP & 8\(\times\)1\(\times\)3 & 2022 & 76 & 362 & 58 & 86.3   &  -\\

 AIM ViT-B-14\cite{yangaim} & IN21K & 8\(\times\)1\(\times\)3 &  624 & 202 & 100 & 14 & 78.8 & - \\
 AIM ViT-L/14~\cite{yangaim} & CLIP & 32\(\times\)1\(\times\)3 &  11208 & 3425 & 341  & 38 & {\bf 87.5} & {\bf 97.7} \\

\midrule
 \textbf{ZeroI2V} ViT-B/16 & IN21K & 8\(\times\)1\(\times\)3 & 422 & 0 & 86 & 0 & 78.6  & -  \\
 \textbf{ZeroI2V} ViT-B/16 & CLIP & 8\(\times\)1\(\times\)3 & 422 & 0 & 86 & 0 & 83.0  & 95.8  \\
 \textbf{ZeroI2V} ViT-B/16 & CLIP & 16\(\times\)1\(\times\)3 & 844  & 0 & 86 & 0 & 83.4 & 96.2 \\  
 \textbf{ZeroI2V} ViT-B/16 & CLIP & 32\(\times\)1\(\times\)3 & 1688 & 0 & 86 & 0 & 83.7 & 96.4 \\ 
 \textbf{ZeroI2V} ViT-L/14 & CLIP & 8\(\times\)1\(\times\)3 & 1946 & 0 & 304 & 0 & 86.3  & 97.4  \\
 \textbf{ZeroI2V} ViT-L/14 & CLIP & 16\(\times\)1\(\times\)3 & 3892 & 0 & 304 & 0 & 86.8  & 97.6 \\ 
 \textbf{ZeroI2V} ViT-L/14 & CLIP & 32\(\times\)1\(\times\)3 & 7783 & 0 & 304 & 0 &  87.2 &  97.6 \\ 
\bottomrule
\end{tabular}
}\vspace{-2mm}

\label{tab:sota_kinetics}
\end{table}
\begin{table}[t] 
    \centering
    \caption{\textbf{Results on Something-Something v2  validation set.} \dag~indicates that the model is pre-trained on both IN21K (except for Uniformer~\cite{uniformer} which uses IN1K) and K400/K600. Other notations are the same as Table \ref{tab:sota_kinetics}.
    }\vspace{-3mm}

    \tablestyle{4.8pt}{1.0}
    \resizebox{\linewidth}{!}{
\begin{tabular}{@{}y{140}x{30}x{30}x{29}x{29}x{29}x{29}x{22}x{22}@{}}
\toprule
Methods & Pretrain & Views & GFLOPs & Extra GFLOPs & Param(M) & New Param(M) & Top-1 & Top-5\\
\toprule
\multicolumn{7}{@{}l}{\it Methods with full fine-tuning} \\

TimeSformer-L~\cite{timesformer} & IN21K & 64\(\times\)3\(\times\)1 & 7140  & - &121 & - &  62.4 & - \\
ViViT-L~\cite{vivit} & K400\dag & 16\(\times\)3\(\times\)4 & 11892  & - & 311 & - & 65.4 & 89.8 \\

MTV-B(\(\uparrow\)320)~\cite{multiview} & K400\dag & 32\(\times\)3\(\times\)4 & 11160  & - & 310 & - & 68.5 & 90.4 \\
VideoSwin-B~\cite{videoswin} & K400\dag & 32\(\times\)3\(\times\)1 & 963  & - &89 & - &  69.6 & 92.7 \\
MViTv2-L(\(\uparrow\)312)~\cite{mvitv2} & K400\dag & 40\(\times\)3\(\times\)1 & 8484  & - & 213 & - & {\bf 73.3} & {\bf 94.1} \\
UniFormer-B~\cite{uniformer} & K600\dag & 32\(\times\)3\(\times\)1 & 777  & - &50 & - &  71.2 & 92.8 \\
ViT-L/14~\cite{vit}  & CLIP & 8\(\times\)3\(\times\)1 & 1946 & 0 & 304 & 0 &  48.7 & 77.5 \\

ILA ViT-L/14~\cite{ila} & CLIP & 8\(\times\)3\(\times\)4 & 10884  & 3100 & 529 & 225  &  67.8 & 90.5 \\
\midrule
\multicolumn{7}{@{}l}{\it Methods with PETL} \\


  ST-Adapter ViT-B/16~\cite{stadapter} & IN21K & 8\(\times\)3\(\times\)1 & 455 & 33 & 93 &  7 & 62.8 & - \\
  ST-Adapter ViT-B/16~\cite{stadapter} & CLIP & 32\(\times\)3\(\times\)1 & 1955 & 267 & 100 & 14 &  69.5 & 92.6 \\
 EVL ViT-L/14~\cite{evl} & CLIP & 32\(\times\)3\(\times\)1 & 9641 & 1858 & 479 & 175 & 66.7  & - \\

 AIM ViT-B/16 & IN21K & 8\(\times\)3\(\times\)1 & 624 & 202 & 100 &  14  & 62.0 & - \\
 AIM ViT-L/14~\cite{yangaim} & CLIP & 32\(\times\)3\(\times\)1 &  11508 & 3725 & 354 & 50 & {70.6} & {92.7} \\

\midrule
\textbf{ZeroI2V} ViT-B/16 & IN21K & 8\(\times\)3\(\times\)1 & 422  & 0 & 86 & 0 &  65.3  & -  \\
\textbf{ZeroI2V} ViT-B/16 & CLIP & 8\(\times\)3\(\times\)1 & 422  & 0 & 86 & 0 &  67.7  & 90.8  \\
\textbf{ZeroI2V} ViT-B/16 & CLIP & 16\(\times\)3\(\times\)1 & 844   & 0 & 86 & 0 &  69.4 & 91.7 \\
\textbf{ZeroI2V} ViT-B/16 & CLIP & 32\(\times\)3\(\times\)1 & 1688  & 0 & 86 & 0 & 70.1  & 92.4 \\ 
\textbf{ZeroI2V} ViT-L/14 & CLIP & 8\(\times\)3\(\times\)1 & 1946  & 0 & 304 & 0 & 70.1  & 91.8  \\
\textbf{ZeroI2V} ViT-L/14 & CLIP & 16\(\times\)3\(\times\)1 & 3892  & 0 & 304 & 0 & 71.4  & 93.0 \\ 
\textbf{ZeroI2V} ViT-L/14 & CLIP & 32\(\times\)3\(\times\)1 & 7783  & 0 & 304 & 0  & {\bf 72.2} & {\bf 93.0} \\ 
\bottomrule
\end{tabular}
}\vspace{-3mm}

\label{tab:sota_sthv2}
\end{table}
\begin{table}[h]
    \centering
    \caption{\textbf{Comparing the state-of-the-art video recognition methods on UCF101, HMDB51 and Diving48.} For UCF101 and HMDB51, we test our method and report the 3-split mean Top-1 accuracy for both datasets following ST-Adapter~\cite{stadapter}. And for Diving48, we test our method with 1 temporal clip following AIM~\cite{yangaim}.}
    \vspace{-2mm}
    \label{tab:ucfhmdbdiving48}
    \resizebox{0.85\linewidth}{!}{
    \begin{tabular}{lcccc}
        \toprule
        Method & Pretrain & ~UCF101~ & ~HMDB51~ & ~Diving48~ \\
        \midrule
        \multicolumn{4}{@{}l}{\it Methods with full fine-tuning} \\

        I3D~\cite{i3dk400} & ImageNet+K400 & 95.6 & 74.8 & - \\
        S3D~\cite{s3d} & ImageNet+K400 & 96.8 & 75.9 & - \\
        SlowOnly-8x8-R101~\cite{slowfast} & Kinetics+OmniSource & 97.3 & 79.0 & - \\
        VideoPrompt~\cite{videoprompt} & CLIP & 93.6 & 66.4 & - \\
        TimeSformer-L~\cite{timesformer} & IN21K & - & - & 81.0 \\
        VideoSwin-B~\cite{videoswin} & IN21K & - & - & 81.9 \\
        \midrule
        \multicolumn{4}{@{}l}{\it Methods with PETL} \\
        AIM ViT-B/16~\cite{yangaim} & CLIP & - & - & 88.9 \\
        AIM ViT-L/14~\cite{yangaim} & CLIP & - & - & 90.6 \\
        ST-Adapter ViT-B/16~\cite{stadapter} & CLIP+K400 & 96.4 & 77.7 & - \\
        ST-Adapter ViT-L/14~\cite{stadapter} & CLIP+K400 & 98.1 & 81.7 & - \\
        \midrule
        \textbf{ZeroI2V} ViT-B/16 & CLIP & 95.6 & 73.7 & 89.7\\
        \textbf{ZeroI2V} ViT-B/16 & CLIP+K400 & 97.7 & 78.5 & -  \\
        \textbf{ZeroI2V} ViT-L/14 & CLIP & 97.8 & 79.9 & \textbf{91.4} \\
        \textbf{ZeroI2V} ViT-L/14 & CLIP+K400 & \textbf{98.6} & \textbf{83.4} & - \\
        \bottomrule
    \end{tabular}
    }\vspace{-2mm}
\end{table}

\subsection{Fully-supervised Experiments}

\noindent\textbf{Results on K400} As shown in Table \ref{tab:sota_kinetics}, our method has significant advantages over traditional full fine-tuning methods, achieving better performance with much lower computational cost. For example, our ZeroI2V ViT-L/14 with an input of 8 frames outperforms MViTv2~\cite{mvitv2}  (86.3 vs 86.1), while requiring more than 20 times fewer GFLOPs (1946 vs 42420). Compared to multi-modal methods such as ActionCLIP~\cite{actionclip} and X-CLIP~\cite{xclip}, which require an additional text branch and fine-tune the entire model end-to-end, our ZeroI2V can achieve comparable performance using only the visual encoder. Moreover, although our proposed ZeroI2V doesn't increase computational or parameter costs during inference compared with the previous PETL method, it can still achieve similar or even better performance. For example, on ViT-B/16, ZeroI2V with an input of 8 frames can surpass ST-Adapter~\cite{stadapter} with an input of 32 frames  
 (83.0 vs 82.7) with much lower GFLOPs (422 vs 1821). On ViT-L/14, ZeroI2V achieves the same performance as EVL~\cite{evl}, which requires an additional 58M parameters. And ZeroI2V achieves comparable performance to AIM~\cite{yangaim} (87.2 vs 87.5) with a nearly 30\% reduction in GFLOPs (7783 vs 11208).

\noindent\textbf{Results on SSv2} As shown in Table \ref{tab:sota_sthv2}, thanks to the effectiveness of STDHA in temporal modeling, our method outperforms most full fine-tuning methods, even though many of them have been pre-trained on the Kinetics dataset. Our ZeroI2V has a significant improvement compared to directly full fine-tuning ViT-L/16 pre-trained with CLIP (70.1 vs 48.7) with the same number of parameters and computation. Compared to other PETL methods, ZeroI2V outperforms ST-Adapter~\cite{stadapter} on ViT-B/16 (70.1 vs 69.5)  with lower GFLOPs (1688 vs 1955). Additionally, ZeroI2V significantly surpasses both EVL~\cite{evl} and AIM~\cite{yangaim} (71.4 vs 66.7, 70.6) on ViT-L/14 with much lower GFLOPs (3892 vs 9641, 11508) and new parameters (0M vs 175M, 50M).

\noindent\textbf{Results on smaller datasets} As shown in Table \ref{tab:ucfhmdbdiving48}, on three relatively small datasets, our method achieves state-of-the-art performance on UCF101, HMDB51, and Diving48. This demonstrates a clear performance advantage over both full-finetuning methods and PETL methods previously.

\noindent\textbf{Results on action detection} In addition to the task of action recognition, to understand the capability of our method in fine-grained spatial understanding, we also evaluate our method on action detection dataset AVA~\cite{ava}. Following the setting of VideoMAE~\cite{videomae}, we evaluate the top 60 common classes using the mean Average Precision (mAP) as the metric under an IoU threshold of 0.5. As shown in Table \ref{tab:ava}, compared to using the original image CLIP features, our ZeroI2V achieved a significant performance improvement (26.4 vs 18.3) with the same number of parameters and computation. It’s noteworthy that our method was not pre-trained on action recognition datasets such as Kinetics. Instead, we directly applied image-to-video transfer on the AVA dataset. Remarkably, our method still managed to achieve performance on par with full-finetuning methods and self-supervised methods that underwent pre-training using the Kinetics dataset, even when using only 8 frames as input. In summary, our ZeroI2V demonstrates outstanding potential in video tasks beyond recognition.


\begin{wraptable}[8]{r}{4.7cm}
    \vspace{-10mm}
    \caption{\textbf{Comparing the state-of-the-art action detection methods on AVA 2.2.}}
    \label{tab:ava}
    \resizebox{\linewidth}{!}{
    \begin{tabular}{lccc}
        \toprule
        Method & Pretrain & Frames & mAP \\
        \midrule
        SlowFast-R101~\cite{slowfast} & K400 & 8 & 23.8 \\
        MViTv2-B~\cite{mvitv2} & K400 & 32 & 28.1  \\
        VideoMAE-B~\cite{videomae} & K400 & 16 &  31.8 \\
        VideoMAE-B~\cite{videomae} & K400 wo/ labels & 16 &  26.7 \\
        \midrule
        CLIP ViT-B/16 & CLIP & 8 & 18.3 \\
        \textbf{ZeroI2V} ViT-B/16 & CLIP & 8 & 26.4  \\
        \bottomrule
    \end{tabular}
    }
    \label{tab:ucf_hmdb}
    \vspace{-15mm}
\end{wraptable}

\subsection{Few-shot Experiments}


\begin{table*}[t]
\caption{\textbf{Comparing the SoTA video recognition methods on the VidTAB~\cite{videoeval}}.}
\vspace{-7mm}
\begin{center}
\resizebox{0.95\linewidth}{!}{
\begin{tabular}{lc|c|cc|cc|cc|c|c}
\toprule
\multirow{2}{*}{} &   & & \multicolumn{2}{c}{\textbf{Action}} & \multicolumn{2}{c}{\textbf{Science}} & \multicolumn{2}{c}{\textbf{Safety}} & \multicolumn{1}{c}{\textbf{Quality}}  & \multicolumn{1}{c}{\textbf{ Emotion}}\\
 & \rotatebox{90}{\# Pretrain Data} & \rotatebox{90}{\textbf{Average}} & \rotatebox{90}{Dark Scene}  & \rotatebox{90}{Long Video} & \rotatebox{90}{Medical Surgery} & \rotatebox{90}{Animal Behavior} & \rotatebox{90}{Harmful Content} & \rotatebox{90}{Fake Face} & \rotatebox{90}{Quality Assess}  & \rotatebox{90}{Emotion Analysis} \\
 
\hline

CLIP-L~\cite{clip} & CLIP & 42.8 & 31.2& 38.0& 32.3& 36.3& 50.3& 58.5& 67.7& 28.1  \\
ViCLIP-L~\cite{internvid} & CLIP+InternVid200M & 42.7 & 36.7& 43.9& 30.2& 36.8& 46.9& 54.8& 65.4& 27.2  \\

ST-Adapter-CLIP-L~\cite{stadapter} & CLIP & 46.9 & 43.0& 45.0& 31.2& 39.4& 49.4& 64.9& 72.3& 29.9  \\
ZeroI2V-CLIP-L & CLIP & 46.5 & 41.3& 46.8& 31.2& 39.3& 47.2& 64.6& 70.6& 30.6  \\

\bottomrule
\end{tabular}
}
\vspace{-6mm}
\end{center}

\label{tab:vidtab_result}
\end{table*}

To demonstrate the adaptation capability of our method in few-shot scenarios, we conduct experiments on the Video Task Adaptation Benchmark (VidTAB). As show in Table~\ref{tab:vidtab_result} The results show that our method can effectively enhance the adaptation of the image model to video tasks using only a few samples. Compared to ST-Adapter~\cite{stadapter}, our approach achieves comparable results while enjoying the advantage of parameter and inference efficiency.

\subsection{Efficiency analysis}

\begin{table}[t]
    \centering
    \caption{\textbf{Inference latency and throughput}. All results are obtained
     using the same V100-32G with PyTorch-builtin mixed precision, using a batch size of 1 to measure latency and the optimal possible batch size to measure throughput before out of memory.}
     \vspace{-3.5mm}
     \resizebox{0.9\linewidth}{!}{
    \begin{tabular}{lcccccc}
        \toprule
        Model & ~Views~  & ~GFLOPs~  & ~Latency (ms)~ & ~Throughput (V/s)~  & ~K400 (Top-1)~ & ~SSv2 (Top-1)~ \\
        \midrule
        Uniformer-B~\cite{uniformer} & 32\(\times\)4 & 1036 & 245.38 & 4.24 & 82.9 & - \\
        EVL ViT-B/16~\cite{evl} & 8\(\times\)3 &  454 & 53.87 & 24.04 & 82.9 & 61.0 \\
        ViT-B/16~\cite{vit} &	8\(\times\)3 	& 422 & 28.72	& 40.08  & 81.0 & 44.0\\
       \textbf{ZeroI2V} ViT-B/16 	& 8\(\times\)3 & 422	& 28.89	& \textbf{40.08}  & \textbf{83.0} & \textbf{67.7} \\
        \bottomrule
    \end{tabular}
    }
    \label{tab:speed}
\end{table}
\begin{table}[ht]
    \centering
        \tablestyle{4.8pt}{1}
        
    \caption{\textbf{Comparison of training cost}. Our results are obtained
     using a same V100-32G with PyTorch-builtin mixed precision, following EVL~\cite{evl}. "\dag" indicates that the epoch is estimated based on the batch size and training steps of the original paper. "Memory" refers to the GPU memory usage when the batch size is 8.}
     \vspace{-3.5mm}
     \resizebox{0.95\textwidth}{!}{
    \begin{tabular}{lcccccc}
        \toprule
        Model (Frames)  & Dataset & \begin{tabular}{c}Training\\Epochs\end{tabular} & \begin{tabular}{c}Training\\GPU Hours\end{tabular}  & Tunable Param (M)  & Memory (G) & Top-1  \\
        \midrule
        Uniformer-B~\cite{uniformer} (32)  & K400  & 110 & 5000 \(\times\) V100 & 50 & - & 82.9 \\
        ActionCLIP ViT-B/16~\cite{actionclip} (16)  & K400 & 50 & 480 \(\times\) RTX3090 & 142 &  - & 82.6 \\
        \midrule
        \multirow{2}{*}{EVL ViT-B/16~\cite{evl} (8)}  & K400 & 53\dag  & 60 \(\times\) V100  & 29  &  2.2 & 82.9 \\
         & SSv2 & 46\dag  & 75 \(\times\) V100  & 98 &  5.6 & 61.0 \\
        \midrule
        \multirow{2}{*}{ST-Adapter ViT-B/16~\cite{stadapter} (8)}
         & K400 & 11\dag & 23 \(\times\) V100  & 7 & 6.9 & 82.0 \\
        & SSv2 & 38\dag & 60 \(\times\) V100  & 14 & 7.6 & 67.1 \\
         \midrule
        \multirow{2}{*}{AIM ViT-B/16~\cite{yangaim} (8)}  & K400 & 30 & 120 \(\times\) V100  & 11 & 8.7 &  83.9 \\
        & SSv2 & 50 & 150 \(\times\) V100  & 14 & 9.0 &  66.4 \\
         \midrule
       \multirow{2}{*}{\textbf{ZeroI2V} ViT-B/16 (8)} & K400  & 40 & 100 \(\times\) V100	& 14 & 7.6 & 83.0 \\
       & SSv2  & 50 & 90 \(\times\) V100	& 14 & 7.6 & 67.3 \\
        \bottomrule
    \end{tabular} }
    \vspace{-5mm}
    \label{tab:train_cost}
\end{table}
\noindent\textbf{Comparison of inference efficiency} We compared the inference efficiency of our method with other methods on the same hardware device. As shown in Table \ref{tab:speed}, under comparable accuracy, the throughput of our method is 10 times that of Uniformer~\cite{uniformer}, Compared to the original ViT-B, our method introduces negligible additional latency during inference while achieving superior performance. In comparison with EVL~\cite{evl}, it can also be seen that the impact of the additional computational module on the actual runtime latency (28.89 ms vs 53.87 ms) is greater than that reflected by GFLOPs (422 vs 454).

\noindent\textbf{Comparison of training cost} We compared the training cost of our method with previous methods in Table \ref{tab:train_cost}. It can be seen that compared to previous full fine-tuning methods such as Uniformer~\cite{uniformer} and ActionCLIP~\cite{actionclip}, our method significantly reduces training cost. Compared to the previous PETL method, our method does not have a significant advantage in training efficiency due to the use of dense adapters. EVL~\cite{evl}, which does not need to insert adapters into the frozen backbone, avoids some of the cost of backpropagation and therefore has lower memory usage. ST-Adapter~\cite{stadapter}, due to its fewer trainable parameters, has a faster convergence speed, but its memory usage is close to our method. Nonetheless, in contrast to AIM~\cite{yangaim} that imposes an additional computational burden for temporal modeling, our STDHA method, which does not introduce extra learnable parameters, ensures that ZeroI2V maintains superior training efficiency. We believe that it is worthwhile and acceptable to exchange some training costs for a reduction in inference costs. We will also try to further reduce training costs by improving training and adaptation strategies in the future.

\section{Conclusions}
\label{sec:Conclusions}

In this work, we present a new approach for {\em parameter-efficient image-to-video transfer learning}, called ZeroI2V. By fully leveraging the powerful representational capabilities of pre-trained image models, our approach enables image transformers to perform video tasks without introducing extra costs during inferences. Our proposed STDHA achieves efficient spatial-temporal modeling at zero extra computation and parameters. In addition, through structural reparameterization and full adaptation strategies, we successfully use a linear structure to achieve zero extra inference cost image-to-video adaptation for the first time. ZeroI2V shows strong performance compared to previous full fine-tuning and PETL methods on five widely used action recognition benchmarks while maintaining parameter and inference efficiency. Due to the simplicity and versatility of our method, we believe it can be easily extended to other video tasks and even multi-modal understanding tasks. We will further investigate this direction in future work.

\paragraph {\bf Acknowledgements.} {This work is supported by the National Key R$\&$D Program of China (No. 2022ZD0160900), the National Natural Science Foundation of China (No. 62076119, No. 61921006), the Fundamental Research Funds for the Central Universities (No. 020214380119), and the Collaborative Innovation Center of Novel Software Technology and Industrialization.}

%
%
\bibliographystyle{splncs04}
\bibliography{main}

\newpage

\section{Appendix}

In this appendix, we provide more details of ZeroI2V from the following aspects:
\begin{itemize}

    \item Implementation details of our method are in \S~\ref{sec:appendix_a}.
    \item Experimental results with other pre-trained weights and backbone architectures can be found in \S~\ref{sec:appendix_b}.
    \item Visualization of our proposed Spatial-Temporal Dual-Headed Attention (STDHA) is in \S~\ref{sec:vis}.
    \item Limitations and societal impact are in \S~\ref{sec:impact}
    \item License of the datasets and pre-trained models are in \S~\ref{sec:license}
\end{itemize}

\subsection{Implementation details of our method}
\label{sec:appendix_a}

\subsubsection{Model Details}
\begin{table}[ht]
\centering
\caption{\textbf{Model details.}  We use a multiset to represent the time offsets of different heads (\eg, "\(1 \cdot 2\)" means that there are 2 heads with \(\Delta t =1\)). When \(\Delta t = 0\), it represents a spatial head."Temporal RF" refers to temporal receptive field of a single STDHA. "Num. Adapters" refers to the number of linear adapters per ViT block.} 
\subfloat[Model details for Kinetics400.]{%
\tablestyle{4.8pt}{1.1}
\label{tab:models_details_k400}
\begin{tabular}{@{}x{50}x{30}y{160}x{30}x{30}@{}}
    \toprule
    Backbone & Frames & \(\Delta t\) of heads & Temporal RF &  Num. Adapters \\
    \midrule
    \multirow{3}{10mm}{ViT-B (\(h\)=12)} & 8 &
    \(\{1\cdot1, -1\cdot1, 0\cdot10\} \) & 3    &  4    \\
  & 16 & \(\{1\cdot1, -1\cdot1,  2\cdot 1,0\cdot9\} \)& 4 & 4    \\
  & 32 & \(\{1\cdot1, -1\cdot1, 2\cdot1, -2\cdot1, 3\cdot1, 0\cdot7\} \) & 6 & 4   \\
  \midrule
\multirow{3}{10 mm}{ViT-L (\(h\)=16)}              &  8 &
    \(\{1\cdot2, -1\cdot2, 0\cdot12\} \) & 3  & 4   \\
  & 16 & \(\{1\cdot2, -1\cdot2, 2\cdot1, 0\cdot11\} \) & 4  & 4    \\
  & 32 & \(\{1\cdot2, -1\cdot2, 2\cdot1, -2\cdot1, 3\cdot1, 0\cdot9\} \) & 6  &  4   \\
  \bottomrule
  \end{tabular}}\hfill
\subfloat[Model details for Something-Something v2.]{%
    \tablestyle{4.8pt}{1.1}
    \label{tab:models_details_sthv2}
       \begin{tabular}{@{}x{50}x{30}y{160}x{30}x{30}@{}}
    \toprule
    Backbone & Frames & \(\Delta t\) of heads & Temporal RF &  Num. Adapters \\
    \midrule
    \multirow{3}{10mm}{ViT-B (\(h\)=12)} & 8 &
    \(\{1\cdot1, -1\cdot1, 0\cdot10\} \) & 3    &  6   \\
  & 16 & \(\{1\cdot1, -1\cdot1, 2\cdot1, -2\cdot1, 0\cdot8\} \)& 5 & 4    \\
  & 32 & \(\{1\cdot1, -1\cdot1, 2\cdot1, -2\cdot1, 3\cdot1, 0\cdot7\} \) & 6 & 4   \\
  \midrule
\multirow{3}{10 mm}{ViT-L (\(h\)=16)}              &  8 &
    \(\{1\cdot2, -1\cdot2, 0\cdot12\} \) & 3  & 4   \\
  & 16 & \(\{1\cdot2, -1\cdot2, 2\cdot2, -2\cdot2, 0\cdot8\} \) & 5  & 4    \\
  & 32 & \(\{1\cdot2, -1\cdot2, 2\cdot1, -2\cdot1, 3\cdot1, 0\cdot9\} \) & 6  &  4   \\
\midrule
\multirow{4}{12 mm}{Swin-B (\(h\) = 4, 8, 16 ,32)} & \multirow{4}{3 mm}{32} &  Stage 1: \(\{1\cdot1, 0\cdot3\} \) & 2 & \multirow{4}{1.5 mm}{4}\\ 
   & & Stage 2: \(\{1\cdot1, -1\cdot1, 0\cdot6\} \) & 3 & \\
   & & Stage 3: \(\{1\cdot1,-1\cdot1,2\cdot1,-2\cdot1, 0\cdot12\} \) & 5 & \\
   & & Stage 4: \(\{1\cdot2,-1\cdot2,2\cdot2,-2\cdot2, 0\cdot24\} \) &  5 &     \\
  \bottomrule
  \end{tabular}}
      \hfill
\end{table}


Our model details are shown in the Table \ref{tab:models_details_k400} and Table \ref{tab:models_details_sthv2}. Due to different requirements for spatial modeling and temporal modeling in different datasets, there are slight differences in the specific implementation settings.

\begin{enumerate}
    \item \noindent\textbf{Settings of STDHA} For the settings of STDHA, we allocate 1/6 to 1/4 of the heads for temporal modeling based on the conclusions obtained from previous ablation experiments. For long inputs, we increase the absolute value of \(\Delta t\) to obtain a larger temporal receptive field. When using Swin-B as the backbone, due to its four stages and different numbers of heads in each stage, we simply allocate temporal heads to each stage at a ratio of 1/4. Since its input length is halved after patch embedding (from 32 frames to 16 frames), we set the value of \(\Delta t\) according to the best temporal receptive field of 16 frames, which is 5. Please note that we have not tried other configurations due to time constraints, so there may be better configurations.
    \item \noindent\textbf{Number of adapters} Considering the balance between performance and training cost, we only assign different adapters for each weight for the minimum setting (ViT-B with 8-frame input) on the SSv2 dataset, which requires 6 adapters. For all other settings, we only use 4 adapters. And the bottleneck ratios of all adapters are set to 0.25.
\end{enumerate}

\subsubsection{Training Details}

\begin{table}[h]
    \centering
    \caption{{\bf Training details of our method.}}
    \resizebox{\linewidth}{!}{
    \begin{tabular}{lcc}
        \toprule
        dataset  & K400 & SSv2 \\
        \midrule
    \multicolumn{3}{@{}l}{\it Optimization settings} \\
        optimizer & \multicolumn{2}{c}{AdamW, learning rate=3e-4, weight decay=5e-2} \\
        batch size & \multicolumn{2}{c}{64} \\
        training epochs & 40   &  50 \\
         \midrule
         \multicolumn{3}{@{}l}{\it Sampling settings} \\
        crop size & \multicolumn{2}{c}{224} \\
       
        frame sampling rate &
  
        \begin{tabular}{c}16 (for 8-frame input)\\8 (for 16-frame input)\\4 (for 32-frame input)\end{tabular}
         &  uniformly sample as TSN~\cite{tsn}   \\
        
         num. testing views & 3 temporal $\times$ 1 spatial & 1 temporal $\times$ 3 spatial \\
        \midrule
         \multicolumn{3}{@{}l}{\it Data augmentation settings} \\
     RandAugment \cite{randaugment} & \multicolumn{2}{c}{m=7, n=4}  \\
        flip & \multicolumn{2}{c}{0.5} \\
        
        Random erasing \cite{rande} & - & 0.25 \\
        label smoothing & - & 0.1 \\

        \bottomrule
    \end{tabular}
    }
    \label{tab:train_details}
\end{table}

As shown in Table \ref{tab:train_details}, our training strategy is similar to the previous methods~\cite{stadapter, yangaim}. Considering that SSv2 requires stronger temporal modeling ability, we used a stronger data augmentation strategy following \cite{videoswin}. In addition, for the full finetuing experiment using Swin-B as the backbone (Swin-B with STDHA), we use exactly the same training strategy as video swin transformer~\cite{videoswin}.

\noindent\textbf{Implementation details of adaptation strategies} The training configurations used for all the adaptation strategies are summarized as follows:
\begin{itemize}
    \item For the comparison experiment of full finetuning ViT with CLIP pretrained, we use 1/10 of the learning rate to avoid training collapse.
    \item For the comparison experiment of only tuning temporal head, we froze the parameters related to the spatial head (only training the part of the parameters related to the temporal head, in other words, we only trained \(W_{\text{attn}}^{Q^t}, W_{\text{attn}}^{K^t}, W_{\text{attn}}^{V^t} \in \mathbb{R}^{d\times d^t}, W_{\text{attn}}^{O^t} \in \mathbb{R}^{d^t\times d}\), where \(d^t\) is the number of channels of the temporal head).
\end{itemize}

For any settings not explicitly mentioned, we assume they align with the training settings of the Linear Adapter.

\subsection{Additional experimental results}
\label{sec:appendix_b}

\begin{table}[ht] 
    \centering
    \caption{\textbf{Results on K400 and SSv2 validation set with ImageNet21K pretrained.} Views = \#frames \(\times\) \#spatial crops \(\times\) \#temporal clips. “GFLOPs” means \(10^9\) FLOPs, "M" means \(10^6\). “Extra GLOPs” refers to the extra computation added to the original ViT under the same number of views. "New Params" refers to additional parameters 
 during inference besides the parameters of the original ViT backbone and linear classifier. Views for all methods are 8\(\times\)1\(\times\)3 for K400 and 8\(\times\)3\(\times\)1 for SSv2}
    \tablestyle{4.8pt}{1.1}
\resizebox{\linewidth}{!}{
\begin{tabular}{@{}y{130}x{30}x{30}x{30}x{30}x{30}x{30}x{25}x{25}@{}}
\toprule
Methods & Pretrain & GFLOPs & Extra GFLOPs  & Param(M) & New Param(M) & K400 Top-1 & SSv2 Top-1 \\
\midrule
\multicolumn{7}{@{}l}{\it Methods with full fine-tuning} \\
TimeSformer~\cite{timesformer} & IN21K &  590 & - & 121 & - & 78.0  & 59.5 \\
X-ViT~\cite{xvit}  & IN21K &  425 & - & 92 & - & 78.5  & 64.4 \\
\midrule
\multicolumn{7}{@{}l}{\it Methods with PETL \& ViT-B/16 } \\
EVL~\cite{evl} & IN21K &  454 & 32 & 115 & 29 & 75.4  & - \\
 ST-Adapter~\cite{stadapter} & IN21K &  455  & 33 & 93 & 7 & 76.6  & 62.8 \\
AIM~\cite{yangaim} & IN21K &  624 & 202 & 100 & 14 &  \textbf{78.8} & 62.0 \\

\textbf{ZeroI2V} & IN21K &  422 & 0 & 86 & 0   & 78.6 & \textbf{65.3} \\ 
\bottomrule
\end{tabular}
}
\label{tab:sota_in21k}
\end{table}

\noindent\textbf{Experiments with ImageNet21K pre-trained weights} In order to investigate the adaptability of our method to different pre-trained weights, we conducted experiments using the same model and training settings on ImageNet21K pre-trained weights. The results are shown in Table \ref{tab:sota_in21k}. It can be seen that our method is still very effective under ImageNet21K weights and can surpass previous full fine-tuning methods. Compared to other PETL methods, our method shows stronger robustness. As shown in Figure \ref{fig:fig1}, when using ImageNet21K pre-trained weights, the advantage of our method over other PETL methods is even greater than when using CLIP pre-trained weights. For example, when using CLIP weights, our method slightly surpasses ST-Adapter~\cite{stadapter} (67.7 vs 67.1), while when using ImageNet21K weights, we have a clear advantage (65.3 vs 62.8).


\begin{table}[ht] 
    \centering

    \caption{\textbf{Results on SSv2 validation set with Swin-B backbone.} K400\dag indicates that the model is pre-trained on both IN21K and K400. The other notations are the same as Table \ref{tab:sota_in21k}}.
    \tablestyle{4.8pt}{1.1}
\resizebox{\linewidth}{!}{
\begin{tabular}{@{}y{120}x{30}x{30}x{30}x{30}x{30}x{28}x{28}@{}}
\toprule
Methods & Pretrain  & Views & GFLOPs  & Param(M) & Tunable Param(M)  & Top-1 & Top-5 \\
\toprule

VideoSwin-B~\cite{videoswin} & K400\dag & 32\(\times\)3\(\times\)1 & 963 & 89 & 89 &69.6  & \textbf{92.7}      \\
PST-B~\cite{tps} & IN21K & 32\(\times\)3\(\times\)1 & 741 & 89 & 89 & 67.4 & 90.9 \\
SIFAR-B~\cite{sifar} & IN21K & 32\(\times\)3\(\times\)1 & 789 & 87 & 87 & 62.6 & 88.5 \\
Swin-B w/ \textbf{STDHA}  & IN21K & 32\(\times\)3\(\times\)1 & 741  & 89 & 89 & \textbf{70.0}  & 92.1 \\
\textbf{ZeroI2V} Swin-B  & IN21K & 32\(\times\)3\(\times\)1  & 741  & 89 &  14  & 67.8 & 91.4 \\ 
\bottomrule
\end{tabular}
}

\label{tab:sota_swin}
\end{table}

\noindent\textbf{Experiments with other backbone} In order to verify the universality of our method, we conducted experiments using the Swin Transformer~\cite{swin} in Table \ref{tab:sota_swin}, which has a hierarchical structure and local window attention. As shown in Table 1, although our method is not specifically designed and adjusted for it, it can still achieve performance comparable or even better than other full fine-tuning methods. To our surprise, when we used a full fine-tuning strategy to train Swin-B using STDHA, we achieved a top-1 accuracy of \textbf{70\%}, which even surpassed VideoSwin-B~\cite{videoswin} pre-trained on the K400 dataset. From this, we can see that our designed STHDA is not only versatile but also has powerful temporal modeling capabilities. In addition, for backbones like Swin Transformer that have more inductive bias and have not been trained on large-scale image-text datasets, full fine-tuning may be able to better utilize the temporal modeling capabilities of STDHA.

\subsection{Visualization}
\label{sec:vis}

\begin{figure}[ht]
    \centering
    \includegraphics[width=1\linewidth]{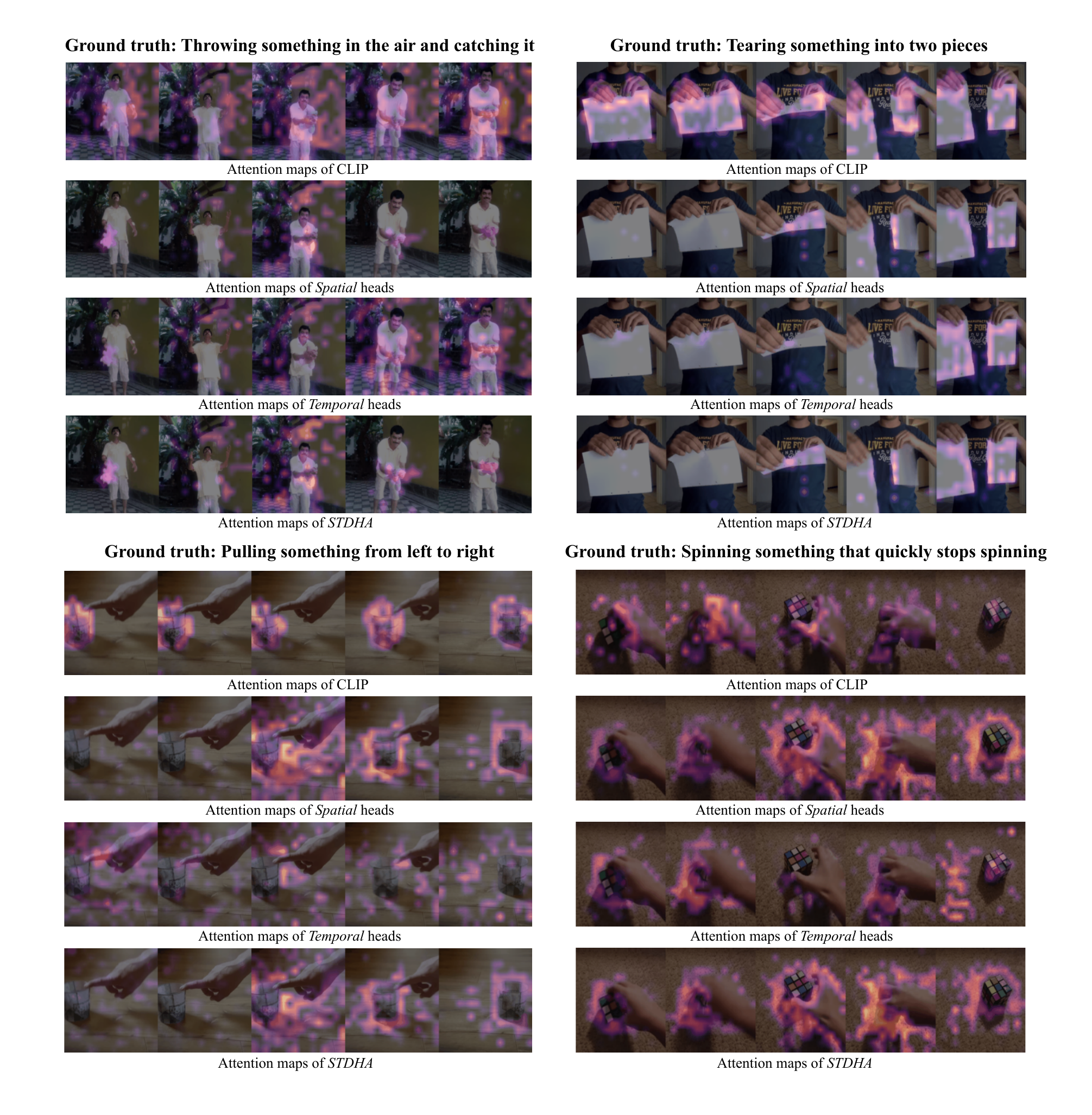}
    \caption{ \textbf{Visualization of attention maps of CLIP, spatial heads, temporal heads and STDHA at the last layer generated by Grad-CAM~\cite{gradcam} on SSv2 validation set.}}
    \label{fig:vis}
    
\end{figure} 

 The motivation behind the design of STDHA is to enable simultaneous spatial and temporal modeling in an independent way before information fusion (ie. decoupling spatial and temporal modeling). In order to more intuitively demonstrate the temporal modeling capabilities of our proposed STDHA, we visualized the attention map of the last layer of the network. As shown in Figure \ref{fig:vis}, note that we visualize the attention map of the last transformer layer in our figure. Due to the temporal receptive field increases with the network depth, both the spatial head and the temporal head of this last layer have a global temporal receptive field about the input frames. But we can still observe that the spatial heads pay more attention to the information of the current frame while the temporal head pays more attention to the information of other frames. We compare the attention maps of STDHA and CLIP, and it can be seen that STDHA pays more attention to the interaction of objects in the video (such as the touch between hands and cups or Rubik’s cubes), while CLIP only focuses on individual objects and does not capture the spatio-temporal dynamics in the video well.

\subsection{Limitations and Societal Impact}
\label{sec:impact}
\noindent\textbf{Limitations} Our method has the following two main limitations:
\begin{itemize}
    \item Although our method is very efficient during inference, the densely inserted linear adapters still need to participate in gradient calculation during training, which brings a non-negligible training cost. This makes our method still have a certain disadvantage in training cost compared to methods that use CLIP as an independent feature extractor (such as EVL~\cite{evl}). In the future, we need to consider more efficient training strategies and improve the structure of linear adapters to address this issue.
    \item Although STDHA has demonstrated powerful temporal modeling capabilities, it still requires consideration of the original number of heads in ViT and manual design of a head relocation strategy. Despite the ablation experiment results showing that our method’s performance is relatively stable across different head relocation strategies, achieving better results still necessitates some manual design. Obtaining optimal head relocation strategies through manual design is obviously challenging. In future work, we aim to investigate methods for automatically designing head relocation strategies.
\end{itemize}

\noindent\textbf{Societal impact} Our ZeroI2V method can apply existing image pre-trained transformers as powerful backbone networks for video tasks such as video classification, spatiotemporal action detection, and video segmentation. Although we do not provide direct applications, it still has the potential to be applied to many scenarios related to video tasks. On the positive side, a powerful video understanding backbone network can improve the performance of downstream tasks and thus enhance efficiency in various scenarios, such as in the fields of smart healthcare and intelligent transportation where video understanding is required. On the other hand, if applied improperly, advanced video networks may also have negative impacts, such as being used in terrorist military activities. Researchers need to carefully consider the potential risks and impacts when applying it to real-world scenarios.

\subsection{License of datasets and pre-trained models}
\label{sec:license}

All the datasets we used are commonly used datasets for academic purpose.  The license of the Kinetics-400\footnote{\url{https://www.deepmind.com/open-source/kinetics}} is CC BY-NC 4.0\footnote{\url{https://creativecommons.org/licenses/by/4.0}}. The license of the Something-Something V2\footnote{\url{https://developer.qualcomm.com/software/ai-datasets/something-something}} is custom. We used the publicly available CLIP pre-trained weights provided by OpenAI\footnote{\url{https://github.com/openai/CLIP}} and the Swin Transformer pre-trained weights provided by Microsoft\footnote{\url{https://github.com/microsoft/Swin-Transformer}}, both of which use the MIT License.

\end{document}